\documentclass[runningheads]{llncs}
\usepackage{graphicx}
\usepackage{booktabs}
\usepackage{multirow}
\usepackage{graphicx}
\usepackage{caption}
\usepackage{subcaption}
\usepackage{hyperref}
\usepackage{textgreek}
\usepackage{makecell}

\begin{document}
\title{Optimising Graph Representation for Hardware Implementation of Graph Convolutional Networks for Event-based Vision
\thanks{The work presented in this paper was supported by: the AGH University of Krakow project no. $16.16.120.773$, the program ''Excellence initiative –- research university'' for the AGH University of Krakow, Polish high-performance computing infrastructure PLGrid (HPC Centers: ACK Cyfronet AGH) -- grant no. $PLG/2023/016130$ and partly by Sorbonne University. Tomasz Kryjak would like to sincerely thank Sorbonne Université for inviting him and funding his stay as a~visiting professor in November 2022.}}


%
\titlerunning{Optimising Graph Representation for GCNs in SoC FPGA}
%
\author{
Kamil Jeziorek\inst{1} \href{https://orcid.org/0000-0001-5446-3682}{\includegraphics[width=12pt]{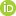}} \and
Piotr Wzorek\inst{1}  \href{https://orcid.org/0000-0003-3885-600X}{\includegraphics[width=12pt]{orcid.png}} \and
Krzysztof Blachut\inst{1} \href{https://orcid.org/0000-0002-1071-335X}{\includegraphics[width=12pt]{orcid.png}} \and
Andrea Pinna  \inst{2}\href{https://orcid.org/0000-0001-5369-0787}{\includegraphics[width=12pt]{orcid.png}} \and
Tomasz Kryjak \inst{1,2}\href{https://orcid.org/0000-0001-6798-4444}{\includegraphics[width=12pt]{orcid.png}}
}


%
\authorrunning{K. Jeziorek et al.}
%
\institute{Embedded Vision Systems Group, Department of Automatic Control and Robotics, AGH University of Krakow, Poland
\email{\{kjeziorek,pwzorek,kblachut,tomasz.kryjak\}@agh.edu.pl}\\
\and
Sorbonne Université, CNRS, LIP6, F-75005 Paris, France\\
\email{andrea.pinna@lip6.fr}
}
\maketitle              
\begin{abstract}
%
Event-based vision is an emerging research field involving processing data generated by Dynamic Vision Sensors (neuromorphic cameras). 
One of the latest proposals in this area are Graph Convolutional Networks (GCNs), which allow to process events in its original sparse form while maintaining high detection and classification performance. 
In this paper, we present the hardware implementation of a~graph generation process from an event camera data stream, taking into account both the advantages and limitations of FPGAs. 
We propose various ways to simplify the graph representation and use scaling and quantisation of values. 
We consider both undirected and directed graphs that enable the use of PointNet convolution. 
The results obtained show that by appropriately modifying the graph representation, it is possible to create a~hardware module for graph generation. 
Moreover, the proposed modifications have no significant impact on object detection performance, only 0.08\% mAP less for the base model and the N-Caltech data set.
Finally, we describe the proposed hardware architecture of the graph generation module.

\end{abstract}
%
%
%
\section{Introduction}



Event cameras, also known as neuromorphic cameras, are an emerging type of a~video sensor.
Unlike standard frame cameras, which read out the brightness values of all pixels simultaneously in a~fixed time interval, event cameras record the brightness changes of individual pixels independently.
This allows correct operation in challenging lightning conditions, such as poor illumination and high dynamic range, and reduces average power consumption.
In addition, information on pixel brightness changes is obtained with a~much higher temporal resolution -- in the order of microseconds vs 16 millisecond for a~typical 60 frame per second frame camera.

One promising application of event cameras is the detection and tracking of moving objects, such as vehicles and pedestrians. 
The use of neuromorphic sensors can significantly improve safety at key moments. 
However, due to their asynchronous nature of operation, event cameras generate an irregular data stream in the form of a~sparse spatio-temporal event cloud, which is difficult to process. 
The most commonly proposed solution is to create pseudo-frames that mimic traditional video frames or to use event-based video frame reconstruction. 
Both approaches produce good results, but lead to the loss of the advantages of high temporal resolution.

Therefore, recent research has focused on processing event data in its original sparse form. 
An example is the use of spiking neural networks (SNNs) \cite{spiking}. Inspired by biology, it is a~promising approach but suffers from a~lack of standardised learning algorithms, making it difficult to apply or achieve high performance in more complex methods.
Another recently proposed solution is the use of Graph Convolutional Networks (GCNs), which allow event data to be processed in the form of a~spatio-temporal point cloud. 
In addition, recent work in this area suggests that such representations can be updated asynchronously \cite{pushing}, \cite{asy_gcn_rec}, \cite{aegnn}. 
However, current computations are mainly performed using high-energy GPUs, and thus similar architectures cannot be used in low-power embedded vision systems.
Furthermore, to the best of our knowledge, current research in data processing with GCNs relies primarily on pregenerated graphs or specifies their construction for a~pre-established structure.

In this paper, we present the hardware implementation of a~graph generation module for event data on SoC FPGAs (System-on-Chip Field Programmable Gate Arrays). Our primary contributions include the development of a~graph representation that takes into account the architectural strengths and limitations of SoC FPGAs, leading to a~reduction in the \textit{number of graph edges} while still achieving detection and classification results comparable to recent methods. Additionally, we introduce the concept of an FPGA hardware module for \textit{graph representation generation}, which produces a~representation ready for processing by the selected PointNet convolution layer \cite{qi2017pointnet}.

The remainder of the paper is organised as follows:
Section \ref{sec:gcn_od} presents the issue of object detection using graph convolutional networks.
Section \ref{sec:sec:our_g} describes the methods we have considered to optimise the graph representation of event data. 
In Section \ref{sec:hw} we present the concept of hardware-based graph generation on FPGAs. 
The article concludes with a~summary \ref{sec:concl} of the results obtained and a~presentation of plans for further work.

\section{Object detection using GCN}
\label{sec:gcn_od}




\subsection{Graph construction}


For an event camera, the event generation process is controlled by a~threshold mechanism. An event is generated when the logarithmic change in light intensity ($\Delta L$) at a~particular pixel ($x_i$, $y_i$) and time $t_i$ exceeds a~fixed threshold value ($C$) for the same pixel at time ${t_i}^*$: $\Delta L(x_i, y_i, t_i) = L(x_i, y_i, t_i) - L(x_i, y_i, t_i^*) \geq p_i C$.



Each generated event includes: the coordinates of the pixel, the time received from the internal camera clock, and the polarisation $p_i$ in the range $\{-1, 1\}$, specifying the direction of the change in light intensity. These four values form a~tuple $e_i = \{x_i, y_i, t_i, p_i\}$, and the events together form a~data sequence $E = \{e_i\}_{i=0}^{N-1}$ ($N$ -- number of events).


The graph structure $\mathcal{G}$ consists of a~set of \textit{vertices} $\mathcal{V}$ and a~set of \textit{edges} $\mathcal{E}$~that reflect the relationships between vertices. 
A commonly accepted method for event-based graph generation is to treat each event as a~vertex $v$ with spatio-temporal position $pos = (x, y, t)$ and attribute $a = (p)$. 
Edges are created based on a~chosen distance metric between vertex positions, most commonly using an Euclidean distance bounded by a~radius of length $R$. A~full description of the graph representations used is given in Section \ref{sec:sec:our_g}.


\subsection{Graph Convolutional Networks}

\subsubsection{Graph convolution.}

Currently, there are many different convolution operations on a~graph, differing in the information used, the functions and the type of data processed. Recent work in this area apply operations such as SplineConv or GCNConv. 
However, the study \cite{jeziorek2023memory} showed that the simpler and memory-efficient PointNetConv operation \cite{qi2017pointnet} can provide comparable results in event data detection and classification problems. 
We therefore decided to use PointNetConv and model presented in \cite{jeziorek2023memory} as a~reference in our work, as it is a~better candidate for embedded implementation.



The graph convolution operation for a~vertex $v_i$ in a~graph $\mathcal{G}$ consists of three steps: a~\textit{message function} that utilises the attributes of a~vertex $a_i$, as well as those of its neighbours $a_j$ and the difference in vertex positions $pos_j - pos_i$, which are transformed by the linear function $h_{\Theta}$. The \textit{aggregation function} then selects the maximum value among all messages for a~given vertex $v_i$. Finally, the \textit{update function} applies an additional linear function $h_{\Theta}$ to the output of the aggregation function to adjust the output size.

These steps are performed for each vertex in the graph $\mathcal{G}$, and at each successive convolution layer, the attribute dimension \(a_i\) of the vertices is modified, with the spatio-temporal position unchanged.
\subsubsection{Graph pooling.}



The pooling process, relevant in the context of GCNs, involves the selection of representative vertices or the aggregation of information within a~group or cluster. Selection criteria may include the relevance of the vertex, the maximum or average value for all vertices in the group. The result of this process is a~new vertex that represents the entire group.

In the context of GCNs, the pooling process enables the creation of denser representations, which translates into a~reduction of computational complexity in subsequent layers. This encourages more efficient processing of large-scale data, while preserving essential features and smooth information flow. It is worth noting that this operation also enables the use of fully connected layers data classification.

\subsection{Related work}


Due to the lack of work that focuses on combining the potential of GCNs, event-based data and hardware implementation on FPGAs, we start the review with methods that use GCNs to process event-based data. We then move on to analyse work that focuses on using FPGAs to accelerate GCNs.

\subsubsection{Object detection using GCN for event data.}





Initially, GCNs were used to process event data because they allowed asynchronous processing at the individual event level.
One of the first work \cite{sekikawa2019eventnet} proposed an EventNet architecture that processed events in a~recursive manner using temporal encoding.
Subsequent work further developed this area, enabling object recognition based on event data. In the work \cite{gcn_class}, residual graph blocks were used, while the work \cite{asy_gcn_rec} introduced asynchronous graph construction and update using slide convolution.
Recent research in this area \cite{pushing}, \cite{aegnn} has shown that object detection is also possible using GCNs. Asynchronicity was achieved by updating only neighbouring vertices in each successive layer, reducing the computational cost and latency of processing each event.
A different approach, compared to previous works, was presented in \cite{jeziorek2023memory}. The focus was on developing memory-efficient GCNs for event data processing, utilising basic convolutional layers tailored for point-cloud data processing.

Despite these advances, all of the work mentioned is mainly based on high-energy platforms such as GPUs. This makes it impossible to develop energy-efficient detection systems using event cameras.

\subsubsection{Graph acceleration on FPGA.}



Implementations of GCNs on FPGAs primarily concentrate on hardware acceleration of the networks themselves. The paper \cite{tao2022lw} introduces a~lightweight GCN accelerator on an FPGA, while \cite{abi2022gengnn} presents an accelerator tailored to various types of convolutional layers. Additionally, the work \cite{engelhardt2019gravf} explores network acceleration on multi-FPGA platforms.

It is worth noting, however, that much of the existing work is centred on processing established graphs or generating graphs based on predefined structures, such as the outputs of accelerator sensors \cite{neu2023real}. In contrast, when dealing with event data characterised by a~random spatio-temporal nature, the required approach differs. It this work, we try to address the latter issue.

\section{The proposed graph representation optimisation}
\label{sec:sec:our_g}

\subsection{Graph representations}



A~variety of graph structure representation methods can be found in the literature, such as Compressed Sparse Row (CSR) and Compressed Sparse Column (CSC). Nevertheless, the most popular format is the Coordinate (COO) format.

In COO format, the edges are represented by vectors containing source and destination vertex indices. In GCNs implementations using off-the-shelf libraries, such as PyTorch Geometric, as demonstrated in works \cite{pushing}, \cite{jeziorek2023memory}, \cite{aegnn}, this representation format is often used because of its simplicity.

Since the number of potentially generated edges linked to an event can vary significantly, initialising sufficient memory to process and store the data in an~FPGA is impossible. To overcome this hardware limit, we propose our own modified representation.

\subsection{Optimisation}



First, it is crucial to determine whether the graph will consist of directed or undirected edges. In our approach, we establish that with the arrival of each new vertex, previous events are directed towards this new event, indicating that the graph is directed based on time. This procedure enables us to apply updates solely to the newly examined vertex during the convolution operation.

As we intend to use the PointNet convolution in our work, it becomes necessary to determine the attributes of a~vertex, its neighbours, and the positional difference between vertices connected by edges. To achieve this, for each vertex, we record its own position $pos = (x, y, t)$, the attribute $a = (p)$ (polarity), and generate edges that specify the connected vertices. For this purpose, we introduce three optimisation methods. 

\subsubsection{Normalisation.}


The locations values of the incoming events depend on the matrix resolution, while the time is represented in microseconds. In existing works \cite{pushing}, \cite{jeziorek2023memory}, \cite{aegnn}, the normalisation only applies to time, where it is multiplied by some factor to make it close to the dimension of the $x$ and $y$ values. Instead, we propose normalising the spatial and temporal dimensions to a~uniform range ($SIZE$). In addition, we quantise the time after normalisation to an integer value. This results in a~spatio-temporal representation where event values are in the range $(0, 0, 0) - (SIZE, SIZE, SIZE)$.

\subsubsection{Neighbour matrix.}





In existing solutions, edge generation requires searching for all vertices or a~relevant subgroup to identify potential neighbours.
Implementing this effectively in an FPGA is difficult due to the requirement of sorting a~large number of vertices.
To address this problem, we propose a~two-dimensional neighbour matrix (\emph{NM}) for edge generation, which has dimensions of $SIZE \times SIZE$ after normalisation. 
When a~new event with position $(x, y, t)$ occurs, we check its local neighbourhood defined by the search radius $R$ in \emph{NM}.
This approach allows us to search for neighbours only in the group of nearest values, while reducing the number of potentially generated edges.
We also avoid the situation where one vertex is connected by edges to several vertices with the same spatial position $(x, y)$ but different times $t$. Only the most recent vertex from a~given pixel is considered.

\subsubsection{Unique events.}


If, for a~given pixel $(x, y)$ in the neighbour matrix, a~vertex with the time $t$ already exists, we treat the appearance of an event with the same time as a~duplicate and do not include it in the edge generation process.\\


By using these methods, the maximum number of potential vertices in the entire graph, which depends on the $SIZE$ parameter, can be determined. The number of generated edges depends on the search context, defined by the radius $R$. 
Due to the use of time-directed edges, vertices with generated edges can be stored in external memory, eliminating the need for their later updates. 

\begin{table}[!tbp]
\caption{Average number of nodes and edges for data normalised to a~given graph size. Methods selected for hardware implementation are in \textbf{bold}.}
\centering
\resizebox{0.65\textwidth}{!}{%
\begin{tabular}{@{}cccccc|cc@{}}
\toprule
\multirow{2}{*}{Graph size} & \multicolumn{3}{c|}{Event Preprocessing} & \multicolumn{2}{c|}{Edge Generation} & \multirowcell{2}[-2pt]{Average \\ Nodes} & \multirowcell{2}[-2pt]{Average \\ Edges} \\ \cmidrule(lr){2-6}
 & Without & Random & Unique & Radius & NM &  &  \\ \midrule
\multirow{6}{*}{128} & x &  &  & x &  & \multirow{2}{*}{56,940} & 836,304 \\
 & x &  &  &  & x &  & 309,711 \\
 &  & x &  & x &  & \multirow{2}{*}{25,000} & 173,868 \\
 &  & x &  &  & x &  & 70,896 \\
 &  &  & x & x &  & \multirow{2}{*}{\textbf{50,498}} & 623,078 \\
 &  &  & x &  & x &  & \textbf{262,935} \\ \midrule
\multirow{6}{*}{256} & x &  &  & x &  & \multirow{2}{*}{56,940} & 138,757 \\
 & x &  &  &  & x &  & 67,043 \\
 &  & x &  & x &  & \multirow{2}{*}{25,000} & 28,892 \\
 &  & x &  &  & x &  & 14,051 \\
 &  &  & x & x &  & \multirow{2}{*}{\textbf{55,925}} & 131,648 \\
 &  &  & x &  & x &  & \textbf{65,214} \\ \bottomrule
\end{tabular}%
}
\label{tab:average-128}
\vspace{-5mm}
\end{table}

\subsection{Ablation studies}



To examine the impact of the optimisations considered, we performed ablation studies on the N-Caltech101 dataset \footnote{The N-Caltech101 dataset is an event-based version of the original Caltech101 dataset. It consists of 8246 samples distributed across 100 classes, with a~maximum resolution of $240 \times 180$ pixels.}. 
From each sample, we extracted a~50 ms time window with the highest event density. 
The ablation studies were divided into two main components: the effect of the methods on the size of the generated graph and the training results of the model considered.
For each of these elements, we distinguished three subgroups of influencing methods: value normalisation, event preprocessing, and edge generation method.

The methods we presented were compared with the solutions proposed in \cite{aegnn}, in which the number of events is sampled at $random$ to a~fixed value (we selected the number to be 25,000) and using the $radius$ method, in which up to 32 neighbours are searched within a~distance $R$ (we selected $R$ to be equal to 3), considering all vertices in the graph. We also compared the methods $without$ taking into account the preprocessing of events.

\subsubsection{Impact of the methods on the size of the generated graph.}


We noticed the biggest difference in the number of generated edges between $radius$ and the $NM$ (neighbour matrix) methods (Table \ref{tab:average-128}). 
This is mainly due to the fact that the $radius$ function generates undirected edges. Additionally, when using $NM$, we avoid edge multiplications within the same pixel that occur in a~short time interval.
Normalisation to 128 means the reduction of around 12\% of the vertices, while for normalisation to 256 the reduction is only 2\%. 
A~separate analysis of the average pixel activity showed that the use of unique values in particular reduced the impact of the occurrence of overly bright pixels.
Although the use of $random$ preprocessing significantly reduces the number of vertices and edges, dropping events to a~fixed value without prior knowledge of how many there will be is impossible.
This shows that by employing our methods, it is feasible to process a~majority of events while generating fewer edges, compared to the method $radius$. Additionally, the uniform structure of the graph allows for the initialisation of an appropriate memory size.

\subsubsection{Evaluation results.}





The best mAP (mean Average Precision) results based on the methods used are presented in Table \ref{tab:accuracy}. 
The model achieving the highest accuracy values utilised normalisation to a~value of 256, along with subsampling and the $radius$ methods. 
Notably, this model outperformed the baseline model, which relied only on temporal normalisation, by achieving a~1.3\% higher mAP value.
Normalisation to a~value of 64 significantly impacted object classification performance, resulting in poorer outcomes.

An analysis of graph directionality revealed that using a~time-directed graph led to only a~marginal 0.8\% loss in mAP compared to an undirected graph. In contrast, the lowest performance was observed for the oppositely timed directed graph, resulting in an almost 2\% decrease compared to the undirected graph.
Furthermore, applying all the optimisation methods, including normalisation to 256, only slightly affects the results with a~0.08\% reduction in the mAP score. 

\begin{table}[!tbp]
\caption{Ablation study for accuracy for different methods.}
\centering
\resizebox{0.7\textwidth}{!}{%
\begin{tabular}{@{}ccccc@{}}
\toprule
\textbf{Method} & \textbf{Normalisation} & \textbf{Pre-processing} & \textbf{Edge Generation} & \textbf{mAP} \\ \midrule
Base model\cite{jeziorek2023memory} & Only time = 100 & Random & Radius & 53.47 \\ \midrule
\multirow{3}{*}{Normalisation value} & 64 & Random & Radius & 43.73 \\
 & 128 & Random & Radius & 51.99 \\
 & 256 & Random & Radius & 54.77 \\ \midrule
\multirow{3}{*}{Direction of edges} & 128 & Random & NM - undirected & 52.20 \\
 & 128 & Random & NM - time directed & 51.40 \\
 & 128 & Random & NM - op. directed & 50.23 \\ \midrule
\multirow{2}{*}{Unique values} & 128 & Unique & NM - time directed & \textbf{51.28} \\
 & 256 & Unique & NM - time directed & \textbf{53.39} \\ \bottomrule
\end{tabular}%
}
\label{tab:accuracy}
\vspace{-5mm}
\end{table}

Table \ref{tab:sota} presents a~comparative analysis of various GCN models in detection task on the N-Caltech101 dataset. Our work uses the model presented in \cite{jeziorek2023memory}. Although its accuracy does not exceed that reported in \cite{pushing}, it significantly reduces the size of the model. The NvS-S model \cite{asy_gcn_rec}, the closest in size to ours, possesses three times as many parameters, but shows an accuracy lower by 18\%. In contrast, the model \cite{aegnn}, which heavily influenced the development of \cite{jeziorek2023memory}, achieves a~6.11\% higher mAP at the cost of being 81 times larger in size.

The presented results demonstrate that, using the described optimisation methods, it is possible to construct a~graph that enables effective object recognition with small accuracy losses. 
Also, our model achieves a~good compromise between accuracy and size, making it well suited for use on embedded platforms.

\begin{table}[!tbp]
\caption{Comparison of different methods for object detection. * was not directly available and estimated based on the publication. Best results in \textbf{bold}.}
\centering
\resizebox{0.35\textwidth}{!}{%
\begin{tabular}{@{}ccc@{}}
\toprule
\textbf{Name} & \textbf{mAP} & \textbf{Parameters {[}M{]}} \\ \midrule
NvS-S \cite{asy_gcn_rec}  & 34.6 & 0.9 \\
AEGNN \cite{aegnn} & 59.5 & 20.4 \\
EAGR-N \cite{pushing} & 62.9 & 2.7* \\
EAGR-L \cite{pushing}  & \textbf{73.2} & 30.7* \\ 
Our & 53.39 & \textbf{0.25} \\ \bottomrule
\end{tabular}%
}
\label{tab:sota}
\vspace{-5mm}
\end{table}


\section{Hardware module concept}
\label{sec:hw}


We developed the optimisation methods for the representation of the event data graph (Section \ref{sec:sec:our_g}) taking into account the hardware constraints of FPGA platforms. In this context, the limitation of embedded memory usage and ensuring a~deterministic number of operations are of key importance.


\subsection{Hardware module requirements}



From a~functional point of view, the graph generation hardware module must meet the throughput and latency requirements to enable real-time event processing, even in highly dynamic scenes. The processing time of a~single event affects the throughput of the system.

The maximum number of events per second, after time quantisation, is $SIZE \times SIZE$ for a~single time unit. 
However, a~situation where the maximum number of events is present is highly unlikely.
A realistic estimate of the required throughput was carried out based on an analysis of the number of events per time unit for the N-Caltech101 dataset.
The maximum average number of events per 1 ms was determined to be approximately 3300, which means an average of $\sim$3.3 events per 1 \textmu s.
The second major constraint on the implementation of the module is the limited amount of internal memory and logic resources of FPGAs.
We address this by quantising input values and optimising memory footprint usage at the cost of a~slight reduction in accuracy.


\subsection{The proposed hardware module}

\begin{figure}[!tbp]
  \centering
  \includegraphics[width=0.7\textwidth]{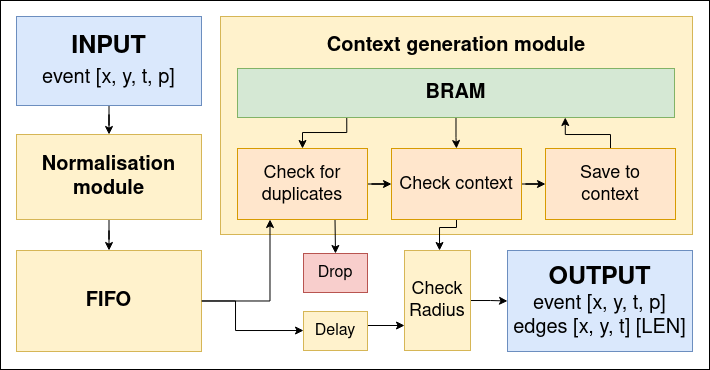}
  \caption{The proposed hardware module scheme}
  \label{fig:fpga}
\end{figure}


Based on the results of the GCN detection study and taking into account the specified system requirements, for the hardware module, we decided on a~graph size of $SIZE=256$, edges directed with time and a~radius of $R=3$.
The module was implemented in the SystemVerilog language, simulated and synthesised using Xilinx Vivado tool for an AMD Xilinx ZCU 104 board with the Zynq UltraScale+ MPSoC chip (XCZU7EV-2FFVC1156).
During testing, we assumed a~module running synchronously with a~clock frequency of at least 250 MHz.

\subsubsection{Input, normalisation and FIFO modules.}

The module's input is a~single event, described by the following signals: $t$ (timestamp), $x$, $y$ (coordinates) and $p$ (polarity).
In addition, the $valid$ flag makes it possible to determine whether an event has been received in a~given clock.


The first operation performed on a~received event is its scaling (for $t$, $x$ and $y$ values) and quantisation to $0-SIZE$ values (Fig. \ref{fig:fpga}, \textit{normalisation} module).
Normalisation performed immediately after receiving the event reduces the use of memory and logic resources.
This operation is executed within a~single clock, so it does not affect system throughput, only latency.


After normalisation, the event can be passed directly to the graph generation module.
However, due to the asynchronous and sparse nature of the event data, we decided to add a~FIFO queue here (Fig. \ref{fig:fpga}, \textit{FIFO}).
In this way, events are accumulated in the queue in moments of increased scene dynamics.
Such a~mechanism allows efficient graph generation even for moments when the number of events per time significantly exceeds the system throughput.
The length of the queue was set considering the availability of memory resources.
After analysing the number of events per time, we assumed its size to be $1024$ elements, which was sufficient for the test sequences used.
The number of bits per memory cell was 25 (3 $\times$ 8-bit coordinates and 1-bit polarity) after normalisation.

\subsubsection{Context generation module.}


The second element of the system is the analysis of the context of the event.
As described in Section \ref{sec:sec:our_g}, we use a~neighbour matrix ($NM$) that contains information about the most recent recorded events.
To store its contents, we used a~two-port BRAM memory of size $SIZE \times SIZE$, with one port for READ and WRITE operations and the other for READ only operations.
To reduce the amount of data stored in the memory, we used the normalised event coordinates as the indexes of particular matrix cells, to which we wrote only the timestamp values.
In this way, we could easily read the events in ($x, y, t$) format from the memory, decoding the coordinates $x$ and $y$, while storing this data in just 8 bits (timestamp only).


For each quantised event read from the FIFO, the content of the memory cell corresponding to its coordinates ($x, y$) in the matrix is checked (Fig. \ref{fig:fpga}, \textit{Check for duplicates} module).
If the timestamp value read from the $NM$ is the same as in analysed event, it is dropped and another one is read from the FIFO queue.



Otherwise, a~context of radius $R=3$ pixels in form of a~square with $2R+1 \times 2R+1$ shape is searched for potential edge candidates.
The stored timestamps are read from the BRAM memory (Fig. \ref{fig:fpga}, \textit{Check context} module).
Each READ operation during the processing of a~single event takes 1 clock cycle (i.e., the latency is 1).
To read the entire context (i.e., 48 memory cells), two ports of the BRAM memory are used, thus reducing the time of this operation twice -- to 24 clock cycles for a~single event.
In practice, it may be sufficient to read fewer neighbours from the $7 \times 7$ context, using one of the approximate representations of a~circle -- then instead of the most time-consuming case of 48 candidates (for a~square), 36 or 24 reads are sufficient, reducing the latency of the module and increasing the number of processable events per time.
However, in our work, we developed the generation module for the worst-case scenario.
It is worth noting that by increasing the width of the BRAM interface (storing multiple context candidates in one memory cell), the bandwidth of the system can be even higher. 



After the entire context is read, a~single WRITE operation to the BRAM memory takes place to update the timestamp value corresponding to the event being processed (Fig. \ref{fig:fpga}, \textit{Save to context} module). At the same time, a~flag is generated indicating that the reading of the context has been completed.

\subsubsection{Output: edges generation.}


The final step is to analyse the context and determine the output signals, which are pairs -- each nondropped input event and a~list of its edges.
A~simple delay line is used to pass the event (in the format $(x, y, t, p)$) to the module output.
Each edge is described with the absolute values of vertices connected by an edge $(x, y, t)$ stored on 24 bits.
By normalising the time and dropping events with the same $(x, y, t)$, an accurate description of a~given vertex and described edge definition allows unambiguous identification of connected vertices.
From such a~generated edge representation, it is thus very easy to switch to the $pos$ and $edges$ structures described in Section \ref{sec:sec:our_g}.


To perform context-based edge generation, a~neighbourhood condition of distance $\leq R$ (in each direction) is checked for each candidate (Fig. \ref{fig:fpga}, \textit{Check radius} module).
We search for the neighbours inside the hemisphere $(x - x_{c})^2 + (y - y_{c})^2 + (t - t_{c})^2 <= R^2$, 
where $(x, y, t)$ are the coordinates of the vertex being processed and $(x_{c}, y_{c}, t_{c})$ are the coordinates of the candidate read from the context.
Resulting pairs (vertex and its edge list) can then be written to external memory 
or passed to the GCN accelerator. 



\subsubsection{Theoretical throughput.}


All logical operations in the proposed module take 1 clock cycle, so the processing of input events can be pipelined.
However, the bottleneck of the solution is the communication with the BRAM memory, where the timestamp values are stored.
A~maximum of 26 clock cycles per single event are required: 25 for READ and 1 for WRITE operation, which was confirmed through simulation.
Based on this information, we determined the theoretical throughput of the system to compare it with the specified requirements.
For a~250 MHz clock, the system is able to accept a~new event once every 104 ns, so over 9.6 events per 1 \textmu s.
The calculated theoretical throughput is almost 3 times higher than the requirements determined based on analysis of the N-Caltech101 dataset. 
Moreover, the use of FIFO module enables a~high performance system for sequences with much higher temporary dynamics in the scene.

\subsubsection{Hardware utilisation.}



The proposed edge generation module was synthesised for an exemplary AMD Xilinx ZCU 104 board with the Zynq UltraScale+ MPSoC chip.
The utilisation of hardware resources is presented in Table \ref{tab:hw}. 

\begin{table}[!tbp]
\caption{Hardware resource utilisation for the proposed edge generation module on an exemplary ZCU 104 platform for graph size of $256 \times 256$.}
\centering
\resizebox{0.35\textwidth}{!}{%
\begin{tabular}{@{}ccc@{}}
\toprule
\textbf{Resource type} & \textbf{Available} & \textbf{Used} \\ \midrule
LUT & 230400 & 5612 (2.4) \\
Flip-Flop & 460800 & 950 (0.2) \\
Block RAM & 312 & 17 (5.5) \\
DSP & 1728 & 189 (10.9) \\ \bottomrule
\end{tabular}%
}
\label{tab:hw}
\vspace{-5mm}
\end{table}







\section{Conclusion}
\label{sec:concl}

In this paper, we presented the concept of a~hardware module that, based on the event stream, can generate a~graph representation that can serve as an input to a~graph convolutional network.
In a~series of experiments, we have shown that our proposed approach using directed (time-based) edges, subsampling and quantisation results in only a~small loss of detection performance -- 0.08\% mAP for the N-Caltech dataset.
The proposed module, in combination with a~GCN accelerator and a~SoC FPGA device, will allow to develop an embedded object detection system based on event data.

Our future objectives include developing a~detection system and exploring several key areas: investigating diverse input data filtering and aggregation techniques, particularly those influenced by SNNs and tackling continuous event stream processing.

\bibliographystyle{splncs04}
\bibliography{mybibliography}






\end{document}